\documentclass{article}

\pdfoutput=1

\usepackage{arxiv}

\usepackage{amsmath,amsfonts,bm}

\def\eqref#1{equation~\ref{#1}}

\def\1{\bm{1}}

\def\vk{{\bm{k}}}

\def\vq{{\bm{q}}}

\def\vv{{\bm{v}}}

\DeclareMathAlphabet{\mathsfit}{\encodingdefault}{\sfdefault}{m}{sl}
\SetMathAlphabet{\mathsfit}{bold}{\encodingdefault}{\sfdefault}{bx}{n}
\newcommand{\tens}[1]{\bm{\mathsfit{#1}}}
\def\tA{{\tens{A}}}

\def\tE{{\tens{E}}}

\def\tG{{\tens{G}}}

\def\tK{{\tens{K}}}

\def\tP{{\tens{P}}}
\def\tQ{{\tens{Q}}}

\def\tV{{\tens{V}}}

\def\gG{{\mathcal{G}}}

\def\sV{{\mathbb{V}}}

\newcommand{\R}{\mathbb{R}}

\usepackage[T1]{fontenc}    
\usepackage{hyperref}       
\usepackage{url}            
\usepackage{booktabs}       
\usepackage{amsfonts}       
\usepackage{nicefrac}       
\usepackage{cleveref}       
\usepackage{graphicx}
\usepackage{natbib}
\usepackage{doi}
\usepackage{caption}
\usepackage{subcaption}
\usepackage{changepage}
\usepackage{relsize}

\graphicspath{{pldrllm_kvgcache_figures/}}

\title{PLDR-LLMs learn a generalizable tensor operator that can replace its own deep neural net at inference}

\date{February 18, 2025}

\newif\ifuniqueAffiliation
\uniqueAffiliationtrue

\ifuniqueAffiliation 
\author{ \hspace{1mm}Burc Gokden \\
	Fromthesky Research Labs LLC\\
	Oregon, USA \\
	\texttt{burc@fromtheskyresearchlabs.com} \\
}
\else
\fi

\renewcommand{\headeright}{preprint}
\renewcommand{\undertitle}{}
\renewcommand{\shorttitle}{PLDR-LLMs learn a generalizable tensor operator that can replace its own neural network}

\begin{document}
\maketitle

\begin{abstract}
We show that Large Language Model from Power Law Decoder Representations (PLDR-LLM) is a foundational model whose deductive outputs are invariant tensors up to a small perturbation. PLDR-LLM learns a singularity condition for the deductive outputs that enable the once-inferred energy-curvature tensor $\tG_{LM}$ to replace the deep neural network of power law graph attention (PLGA) generating the deductive outputs at inference. We demonstrate that a cache for $\tG_{LM}$ (G-cache) and KV-cache can be implemented in a straightforward manner to improve the inference time. The invariance and generalizable nature of deductive outputs is at a very high fidelity where deductive outputs have same RMSE and determinant values up to 15 decimal places after caching, and zero-shot benchmark scores remain unchanged. Ablation studies show that learned deductive outputs have distinct loss and accuracy characteristics from models pretrained with transferred, randomly initialized or identity tensors as a constant tensor operator and an LLM with scaled-dot product attention (SDPA) is a special case of PLDR-LLM where $\tG_{LM}$ is predefined as identity. The observed invariance characteristic introduces a novel asymmetry between training and inference phases with caching. We outline observed common characteristics of the deductive outputs for the learned singularity condition. We provide an implementation of a training and inference framework for PLDR-LLM with KV-cache and G-cache. 
\end{abstract}

\section{Introduction}

Large Language Model from Power Law Decoder Representations (PLDR-LLM) is a novel language model architecture with well-defined deductive and inductive outputs \citep{Gokden2024pldrllm}. It is composed of deep layers of decoders with multi-headed Power Law Graph Attention (PLGA) \citep{Gokden2021, Gokden2019}. The deductive outputs are intended to observe and regularize the model, while the inductive output is the next-token prediction of a language model. PLGA is a series of non-linear and linear transformations that attend to an input sentence that can be considered as a weighted graph $\gG=\left( \sV , E \right)$ where nodes are the tokens densely represented by an N-dimensional embedding space. The PLGA learns a metric tensor $\tA_{LM}$ of the embedding space after applying a custom fully connected layer and iSwiGLU, a positive semi-definite activation function, to the output  $\tA$ of  a deep residual network of gated linear units (GLUs) whose input is a density matrix operator derived from the query. The range and strength of the interactions between each embedding dimension are determined through learned power coefficients $\tP$ that define a potential tensor $\tA_{\textbf{P}}=\tA_{LM}^{\odot\tP}$. Finally, a superposition of these potentials define the energy-curvature tensor $\tG_{LM}$, which represents interaction of each embedding dimension with all other dimensions. The attention $\tE_{LM}$ is then derived by projecting the query and key vectors on $\tG_{LM}$. 

The metric tensor, potential tensor and the energy-curvature tensor are the deductive outputs that were considered in implementation of the PLDR-LLMs \citep{Gokden2024pldrllm} to derive the Directed Acyclic Graph (DAG) loss and use it as a regularizer to modify model characteristics without scaling the model or dataset size. In the study that first introduced the PLDR-LLMs, the focus was on the characterization of model performance with respect to scaling layer depth and model size under the constraint of memory size. It was demonstrated that the PLDR-LLM has comparable performance to reference models (LLMs with Scaled Dot-Product Attention (SDPA)) with similar model size from the literature. The characteristics of $\tA_{LM}$, $\tA_{\textbf{P}}$ and $\tG_{LM}$ at the time of training were explored while evaluating the DAG loss as a metric and regularizer.

In this paper, we investigate the inference characteristics of $\tA_{LM}$, $\tA_{\textbf{P}}$, $\tG_{LM}$ and the direct output of residual network $\tA$ in depth. We report that an efficient implementation PLDR-LLM for inference shows that it is a new kind of foundational model that introduces unique mechanisms and provides a better insight into our understanding of language models in general. We make the following contributions:
\begin{itemize}
	\item We empirically show that the locally defined energy-curvature tensor $\tG_{LM}$, and the set of operators \{$\tA_{\textbf{P}}$, $\tA_{LM}$, $\tA$\} are learned as generalizable tensor operators, such that their generating neural network can be replaced by an input-invariant, generalizable tensor $\tG_{LM}$ after inferring it only once with initial prompt as input.
	 
	\item The learned metric tensor is singular, and the replacement of non-linear transformations by a tensor operator is a result of this learned singularity condition from entire dataset. The deductive outputs exhibit same distribution of values and characteristics to a very high fidelity after removal of the generating network such that the benchmark scores remain unchanged.
	
	\item The above observation also reveals that LLM with SDPA is a special case of PLDR-LLM where the tensor operator $\tG_{LM}$ is identity. We show that PLDR-LLM with a learned tensor operator performs slightly better than an LLM with SDPA under same training conditions.
	
	\item Since $\tG_{LM}$ is an input invariant tensor operator during inference, it can be cached. Implementation of KV-cache at inference becomes straightforward for PLDR-LLM and provides the same benefits as it is for the language model implementations with SDPA.
	
	\item PLDR-LLM introduces a fundamental asymmetry between training and inference phases. $\tG_{LM}$ can replace the deep PLGA net at inference and provide same inductive output up to a small perturbation of its deductive outputs, however training with a PLGA net is not identical to training with a predefined $\tG_{LM}$.
	
	\item An efficient Pytorch implementation of PLDR-LLM for multi-gpu training and inference with KV-cache and G-cache is provided at \url{https://github.com/burcgokden/PLDR-LLM-with-KVG-cache}. 
\end{itemize}

\section{Approach}

The training approach is same as the PLDR-LLMs trained in \citep{Gokden2024pldrllm} and similar to the approaches for training followed in \citep{Radford2019gpt2, Touvron2023llama, Touvron2023llama2}. PLDR-LLMs are trained autoregressively while minimizing the cross-entropy loss (and DAG loss of deductive outputs). We evaluated the pretrained PLDR-LLMs with learnable $\tG_{LM}$ through a PLGA network and with predefined $\tG_{LM}$ on benchmarks for zero-shot performance.

The model parameters of PLDR-LLMs pretrained are shown in table \ref{table1} along with the modified versions of PLDR-LLM that replaces the deep PLGA net (residual gated linear units, application of iSwiGLU, custom linear layer weights and biases, and learnable power coefficients) with a constant $\tG_{LM}$ for ablation studies. PLDRv51 version is the Pytorch implementation of PLDRv5 design \citep{Gokden2024pldrllm} with unused activation and dropout layers removed for simplicity. The number of embedding dimensions per head at each decoder layer was set at $d_{k}=64$. Number of residual layers and number of SwiGLU and LU in each residual layer were set at 8 and 2, respectively. PLDRv51G version does not have trainable PLGA net, instead uses a predefined $\tG_{LM}$ provided as a model hyperparameter during model initialization. PLDRv51Gi version also does not have PLGA net but a predefined $\tG_{LM}$ inferred from an already trained PLDR-LLM of same configuration for input prompt as empty string by generating single token with greedy sampling. It is used only to demonstrate inference by transferring all remaining learned parameters from an already pretrained PLDR-LLM of PLDRv51 type. For the special case where $\tG_{LM}$ is set as identity tensor, PLDRv51G becomes equivalent to an LLM with SDPA.

The models are implemented with the option to enable KV-cache (see, for example, \citep{Shazeer2019onewrite, Liu2024kivi}) and G-cache for faster inference. The KV-cache implementation uses the same approach used in LLMs with SDPA such as GPT and LLAMA. This is possible because $\tG_{LM}$ ($\tA_{LM}$, $\tA$) behaves as an input invariant tensor up to a small perturbation during inference.
 
\textbf{$\tG$-cache.} After processing the prompt as input,  $\tG_{LM}$ is cached once. For the remaining next-token predictions, the neural network that outputs  $\tG_{LM}$ is skipped and cached $\tG_{LM}$ tensor is used as a linear operator.

\textbf{KV-cache.} For the initial prompt, the key, query and value inputs are processed and cached. The output of residual gated linear units $\tA$ is also cached once at this step. For each newly generated token prediction vectors $\vq$, $\vk$, and $\vv$; the single token vector is concatenated to the cached $\tK$ and $\tV$, while $\vq$ propagates as a single token:

For $\tQ, \tK, \tV \in \R^{b \times h \times s \times d_{k}}$ and $\vq, \vk, \vv \in \R^{b \times h \times 1 \times d_{k}}$ after applying linear fully-connected layer and splitting into heads:
\begin{itemize}
	\item Initial caching $\tK$ and $\tV$ of the prompt during first next-token prediction.
	\begin{eqnarray}
		&\tQ&=RotaryEmbedding(\tQ) \\
		&\tK_{cached}&=RotaryEmbedding(\tK) \\
		&\tV_{cached}&=\tV 
	\end{eqnarray}
	\item Use of single token key and value vectors $\vk$ and $\vv$ for subsequent next token predictions.
	\footnote{We keep track of token positions for rotary embedding.}
		\begin{eqnarray}
			&\vk_{update}&=RotaryEmbedding(\vk) \\
			&\vv_{update}&=\vv \\
			&\tK_{cached}&=Concatenate(\tK_{cached}, \vk_{update}) \\
			&\tV_{cached}&=Concatenate(\tV_{cached}, \vv_{update}) 
		\end{eqnarray}
	\item Use of single token query vector $\vq$ for attention without masking:
	\begin{eqnarray}
		&\vq_{update}&=RotaryEmbedding(\vq) \\
		&\tE&=\vq_{update}\tG_{LM}\tK^{T}_{cached} \\
		&\tE_{LM}&=softmax(\tE) \\
		&\vv_{next}&=\tE_{LM}\tV_{cached} 
	\end{eqnarray}
	
\end{itemize}

where $b$, $s$, $h$, and $d_{k}$ are batch size, prompt token length, number of attention heads, and number of embedding dimensions for each head respectively.

The implementation of PLDR-LLM train and inference framework developed for this paper improves upon learnings from  the implementation for \citep{Gokden2024pldrllm}, it is typically faster even without KV-cache and G-cache. For training, we used the fully sharded data parallelism strategy \citep{Zhao2023fsdp}.
\begin{table}
	\caption{Parameters for PLDR-LLMs trained for the experiments and ablation studies. SwiGLU:LU is the layer size for Gated Linear and Linear Units in each residual layer, LR is learning rate, WUP is warm up step size, $d_{ff}$ is the feedforward network layer size at the end of each decoder layer, $\text{\#}ResL/\text{\#}A$ is ratio of total number of parameters in a residual unit to number of entries for $\tA$ for single head at a decoder layer. Data column shows the RefinedWeb data interval used for pretraining.}
	\label{table1}
	\centering
	\resizebox{\textwidth}{!}{
		\begin{tabular}{c c c c c c c c c c c}
			\toprule[1.2pt] 
			Model & \# Layers & \# Heads & $d_{model}$ & $d_{ff}$ & SwiGLU:LU & $\text{\#}ResL/\text{\#}A$ & LR & WUP & $\tG_{LM}$ & Data \\
			\midrule[1.2pt]
			PLDRv51-104M & 7 & 12 & 768 & 2048 & 170:64 & 129.33 & $1.2 \times 10^{-3}$ & 8000 & Trainable & [0, 16M] \\
			\midrule
			PLDRv51-DAG-110M & 5 & 14 & 896 & 2389 & 170:64 & 129.33 & $1.5 \times 10^{-3}$ & 2000 & Trainable & [0, 16M] \\
			\midrule
			PLDRv51-110M-1 & 5 & 14 & 896 & 2389 & 170:64 & 129.33 & $1 \times 10^{-3}$ & 4000 & Trainable & [0, 16M] \\
			\midrule
			PLDRv51-110M-2 & 5 & 14 & 896 & 2389 & 170:64 & 129.33 & $1 \times 10^{-3}$ & 4000 & Trainable & [16M, 32M] \\
			\midrule
			PLDRv51-110M-3 & 5 & 14 & 896 & 2389 & 180:64 & 136.91 & $1 \times 10^{-3}$ & 2000 & Trainable & [0, 16M] \\
			\midrule
			PLDRv51-110M-4 & 5 & 14 & 896 & 2389 & 181:64 & 137.66 & $1.5 \times 10^{-3}$ & 2000 & Trainable & [0, 16M] \\
			\midrule
			PLDRv51-110M-5 & 5 & 14 & 896 & 2389 & 196:64 & 149.03 & $1.5 \times 10^{-3}$ & 2000 & Trainable & [0, 16M] \\
			\midrule[1.2pt]
			PLDRv51Gi-99M & 7 & 12 & 768 & 2048 & NA & NA & NA & NA & PLDRv51-104M & NA \\
			\midrule
			PLDRv51Gi-106M-1 & 5 & 14 & 896 & 2389 & NA & NA & NA & NA & PLDRv51-110M-1 & NA \\
			\midrule
 			PLDRv51Gi-106M-2 & 5 & 14 & 896 & 2389 & NA & NA & NA & NA & PLDRv51-110M-2 & NA \\
 			\midrule
 			PLDRv51Gi-106M-3 & 5 & 14 & 896 & 2389 & NA & NA & NA & NA & PLDRv51-110M-3 & NA \\
 			\midrule
 			PLDRv51Gi-106M-4 & 5 & 14 & 896 & 2389 & NA & NA & NA & NA & PLDRv51-110M-4 & NA \\
 			\midrule
 			PLDRv51Gi-106M-5 & 5 & 14 & 896 & 2389 & NA & NA & NA & NA & PLDRv51-110M-5 & NA \\
 			\midrule
			PLDRv51Gi-DAG-106M & 5 & 14 & 896 & 2389 & NA & NA & NA & NA & PLDRv51-DAG-110M & NA \\
			\midrule[1.2pt]
			PLDRv51G-106M-1 & 5 & 14 & 896 & 2389 & NA & NA & $3 \times 10^{-4}$ & 2000 & PLDRv51-110M-1 & [16M, 32M] \\
			\midrule
			PLDRv51G-106M-2 & 5 & 14 & 896 & 2389 & NA & NA & $3 \times 10^{-4}$ & 2000 & Identity & [16M, 32M] \\
			\midrule
			PLDRv51G-106M-3 & 5 & 14 & 896 & 2389 & NA & NA & $3 \times 10^{-4}$ & 2000 & Random & [16M, 32M] \\
			\bottomrule[1.2pt]
		\end{tabular}
}
\end{table}

\section{Dataset}

Two large sample intervals from the RefinedWeb \citep{Penedo2023falcon} dataset are used to pretrain PLDR-LLMs through $\sim$8B tokens. First sample interval is from first 16M samples, of which a total of 500k batches with batch size of 16 were generated and distributed evenly onto two ranks for pretraining. Second sample interval was between 16M and 32M, and same amount of batches were generated to pretrain additional PLDR-LLMs for transfer learning and ablation studies.

Data preparation follows the same approach that was detailed in \citep{Gokden2024pldrllm}. The context length was set at 1024 tokens. A new SentencePiece unigram tokenizer \citep{Kudo2018sentencepiece, Kudo2018subword} model was trained from RefinedWeb dataset with the same parameters. The preprocessing of samples for batching tokenized text to the context length was optimized to remove occasional padding that may be encountered in a batch.

\section{Experiments}
We conducted a number of experiments to evaluate deductive outputs and to compare benchmark evaluation performance of PLDR-LLMs with different KV-cache and G-cache settings at inference, and with custom initial $\tG_{LM}$ values during training (Table \ref{table1}). These results were also compared to reference LLMs of similar model size reported in the literature.

We pretrained a 7-layer 12-head model (PLDRv51-104M), 5-layer 14-head PLDR-LLMs without (PLDRv51-110M-1, 3, 4 and 5) and with (PLDRv51-DAG-110M) DAG regularization over $\sim$8B tokens obtained from first 16M samples of RefinedWeb dataset. The regularization coefficients for deductive outputs ($\tA_{LM}$, $\tA_{\textbf{P}}$, $\tG_{LM}$) were (0.05, 0.05, 0.05). The coefficients were not optimized for the tokenizer model through a comprehensive parameter search for the best benchmark performance. 

PLDRv51-110M-2 was pretrained on the interval [16M, 32M] from the same dataset without DAG regularization. For comparison, we also evaluated reference models in the literature (GPT2-124M\footnote{https://huggingface.co/openai-community/gpt2} \citep{Radford2019gpt2}, GPT-Neo-125M\footnote{https://huggingface.co/EleutherAI/gpt-neo-125m} \citep{gptneo,Gao2020pile} and Phytia-160M\footnote{https://huggingface.co/EleutherAI/pythia-160m-deduped} \citep{Biderman2023pythia}) in zero-shot setting with their implementation on the Huggingface platform, using EleutherAI Evaluation Harness Suite \citep{evalharness}. The benchmarks were evaluated with KV-cache and G-cache enabled and disabled for PLDR-LLMs with the same evaluation suite.

To observe characteristics of trainable and pre-defined $\tG_{LM}$ values, we ran ablation studies by training PLDR-LLMs in a different section of RefinedWeb dataset using first $\sim$8B tokens from the [16M, 32M] sample interval. For characterization of a model with transfer learning of $\tG_{LM}$, PLDRv51G-106M-1 was pretrained using a $\tG_{LM}$ learned and inferred from PLDRv51-110M-1 pretrained on the data interval [0, 16M]. We also pretrained models where $\tG_{LM}$ is identity (PLDRv51G-106M-2), and a tensor with random values from normal distribution with unit variance and zero mean (PLDRv51G-106M-3) under same training parameters. PLDRv51G-106M-2 is a special case which is equivalent to an LLM with SDPA widely used in the literature. It was possible to train these models at a much lower learning rate of $3 \times 10^{-4}$ and warm up step size of 2000.

PLDR-LLM is sensitive to the choice of SwiGLU:LU ratios which also affect the learning rate and warm up step size. We trained  PLDRv51-110M-3 with a SwiGLU:LU ratio of 180:64 at a learning rate of $1 \times 10^{-3}$ and warm up step size of 2000. PLDRv51-110M-4 and PLDRv51-110M-5 were trained with SwiGLU:LU ratios of 181:64 and 196:64 and a larger learning rate of $1.5 \times 10^{-3}$ for comparison with the base model PLDRv51-110M-1. With base model SwiGLU:LU at 170:64, the ratios were chosen to skew around the $\text{\#}ResL/\text{\#}A$ value of 137\footnote{We choose this value out of convenience, it is widely known as the fine-structure constant ($1/137.036$).}.

PLDRv51Gi-* were used for inference only. These models have PLGA replaced with a predefined $\tG_{LM}$, which was transferred from their respective PLDRv51 type models along with their remaining learnable parameters. The replacement of PLGA with predefined $\tG_{LM}$ reduces the trainable model parameter size to 106M from 110M for the models with 5 layers and 14 heads and to 99M from 104M for the model with 7 layers and 12 heads.

These pretrained models are evaluated on a set of full-size benchmarks (ARC \citep{Clark2018arc}, Hellaswag \citep{Zellers2019hellaswag}, WinoGrande \citep{Keisuke2019winogrande}, TruthfulQA \citep{Lin2021truthfulqa}, OpenBookQA \citep{Mihaylov2018obqa}, PIQA \citep{Bisk2020piqa}, SIQA \citep{Sap2019siqa}) for commonsense reasoning, question answering and language understanding for zero-shot response. For tokenization agnostic scoring, we used byte-length normalized accuracy except for TruthfulQA which uses a custom normalized accuracy for multiple choice, multiple true answers. Following \citep{Gokden2024pldrllm}, two average scores, Avg-1 (without TruthfulQA) and Avg-2 (with TruthfulQA) are reported.

We also evaluated benchmarks with mismatched $\tG_{LM}$ tensors as a negative test. We replaced the actual $\tG_{LM}$ for PLDRv51-110M-1 with identity (PLDRv51-106M-1-NAB1) and a tensor with random values from a normal distribution with unit variance and zero mean (PLDRv51-106M-1-NAB2) at inference. These models with mismatched $\tG_{LM}$ tensors do not generate meaningful continuation from an input prompt, however the models are still partially conditioned with the dataset they are pretrained on. Their effective model size is reduced to 106M.

The models were pretrained on two RTX 4090 GPUs with 24 GB of RAM with the framework developed for this study. Inference was carried on single RTX 4090 GPU.

\section{Results}

\subsection{Evaluation of Deductive Outputs}

The RMSE of difference of deductive outputs between heads at each decoder layer are shown in table \ref{table2} with and without KV-cache and G-cache enabled, and with greedy sampling for 100 tokens. The final RMSE value is calculated across all decoder layers. The RMSE value when caching enabled is same as without caching up to at least 15 decimal digits for most of the models for $\tA_{LM}$, $\tA_{\textbf{P}}$, and $\tG_{LM}$. The perturbation effect is more evident for $\tA$, however this does not reflect on the other deductive outputs derived from it. Deductive outputs derived from empty string fluctuate more for $\tA_{\textbf{P}}$. The perturbation can be modified with DAG loss as PLDRv51-DAG-110M shows the largest deviation with caching among the models. The RMSE of $\tA$ is minimum at $\text{\#}ResL/\text{\#}A=136.91$ among models with same number of decoder layers and attention heads (PLDRv51-110M-1 to 5, PLDRv51-DAG-110M).

The maximum magnitude (absolute value) of determinants derived from deductive output heads are shown in table \ref{table3} up to 15 decimal digits. The determinant of all heads for $\tA$ and $\tA_{LM}$ are zero, hence these deductive outputs are singular for all models trained. We see similar high degree of fidelity to the values without caching for maximum magnitude of determinants of $\tA_{\textbf{P}}$ and $\tG_{LM}$. $\tA_{\textbf{P}}$ and $\tG_{LM}$ typically exhibit very large determinant values on some of the heads. The DAG loss regularization brings the maximum magnitude of determinants observed for $\tA_{\textbf{P}}$ and $\tG_{LM}$ close to zero.

Tables \ref{table2} and \ref{table3} empirically reveal that there are some common characteristics of deductive outputs among models. Combined with observed and a priori known characteristics of row and column values of $\tA$ and $\tA_{LM}$, we can state the following:

\begin{itemize}
	\item $\tA$  has the same set of row values for every head in each layer up to a very small perturbation in their values among rows and different heads within same layer. $\tA$ approximates a matrix-rank 1 tensor and it is singular. It has one non-zero eigenvalue which is approximately the sum of elements in its row. This real eigenvalue represents the spectral radius of $\tA$ for heads at each layer.
	
	\item $\tA_{LM}$ is guaranteed to be a positive-definite tensor for numerical stability\footnote{Through application of iSwiGLU and a small positive bias value \citep{Gokden2021}.}. Since every head is a square matrix $d_{k} \times d_{k}$ of positive real values, and by Perron-Frobenius theorem, they have real eigenvalues which are the spectral radius for each head. As observed empirically, it is also singular for each head.
\end{itemize}

These common characteristics indicate that PLDR-LLM learns a generalizable, non-trivial singularity condition for the deductive outputs from the dataset and under this condition a portion of neural network can be replaced with an invariant tensor operator up to a very small perturbation in deductive outputs.

\begin{table}[!htb]
	\centering
	\caption{RMSE (Root Mean Square Error) of difference between heads at each decoder layer among all layers for the deductive outputs of PLDR-LLMs. The prompt for all models except PLDRv51Gi-* was "Write a letter requesting people use language models responsibly." and the continuation was generated for 100 tokens with greedy sampling (top-k=1). PLDRv51Gi-* deductive outputs are inferred by empty string prompt for single token prediction from their respective PLDRv51 models with greedy sampling. Values are shown up to 15 decimal digits.}
	\label{table2}
	\resizebox{\textwidth}{!}{
		\begin{tabular}{c c c c c c c}
			\toprule[1.2pt]
			Model & \multicolumn{2}{c}{Cache} & \multicolumn{4}{c}{RMSE} \\
			 \cmidrule(lr){2-3} \cmidrule(lr){4-7}
			 & KV & G & $\tA$ & $\tA_{LM}$ & $\tA_{\textbf{P}}$ & $\tG_{LM}$ \\
			\midrule[1.2pt]
			PLDRv51-104M & $\checkmark$ & $\checkmark$  & $1.372936520027679\times 10^{-8}$ & $1.086511760950089 \times 10^{-1}$ & $2.914173364639282$ & $2.760209798812866$ \\
			\midrule
			PLDRv51-104M & $\times$ & $\times$ & $1.320242049018816 \times 10^{-8}$ & $1.086511760950089 \times 10^{-1}$ & $2.914173364639282$ & $2.760209798812866$ \\
			\midrule
			PLDRv51Gi-99M & $\checkmark$ & NA & $6.454119461096752\times 10^{-9}$ & $1.086511760950089 \times 10^{-1}$ & $2.914173603057861$ & $2.760209560394287$ \\
			\midrule[1.2pt]
			PLDRv51-110M-1 & $\checkmark$ & $\checkmark$  & $8.941987705846088 \times 10^{-10}$ & $3.401382640004158 \times 10^{-2}$ & $3.453436136245728$ & $2.910791397094727$ \\
			\midrule
			PLDRv51-110M-1 & $\times$ & $\times$ & $9.683998047904652 \times 10^{-10}$ & $3.401382640004158 \times 10^{-2}$ & $3.453436136245728$ & $2.910791397094727$ \\
			\midrule
			PLDRv51Gi-106M-1 & $\checkmark$ & NA & $1.175712305290233 \times 10^{-9}$ & $3.401382640004158 \times 10^{-2}$ & $3.453436136245728$ & $2.910791397094727$ \\
			\midrule[1.2pt]
			PLDRv51-110M-2 & $\checkmark$ & $\checkmark$ & $1.459935228265152 \times 10^{-10}$ & $1.342141479253769 \times 10^{-1}$ & $3.232312202453613$ & $2.808657884597778$ \\
			\midrule
			PLDRv51-110M-2 & $\times$ & $\times$ & $1.407280958432011 \times 10^{-10}$ & $1.342141479253769 \times 10^{-1}$ & $3.232312202453613$ & $2.808657884597778$ \\
			\midrule
			PLDRv51Gi-106M-2 & $\checkmark$ & NA &$1.325615173186634 \times 10^{-10}$ & $1.342141479253769 \times 10^{-1}$ & $3.232312202453613$ & $2.808657884597778$ \\
			\midrule[1.2pt]
			PLDRv51-110M-3 & $\checkmark$ & $\checkmark$ & $1.234047170006747 \times 10^{-10}$ & $8.118956466205418 \times 10^{-4}$ & $3.703177452087402$ & $2.982294797897339$ \\
			\midrule
			PLDRv51-110M-3 & $\times$ & $\times$ & $9.384986537908091 \times 10^{-11}$ & $8.118956466205418\times10^{-4}$ & $3.703177452087402$ & $2.982294797897339$ \\
			\midrule
			PLDRv51Gi-106M-3 & $\checkmark$ & NA & $1.137204427847927 \times 10^{-10}$ & $8.118956466205418 \times 10^{-4}$ & $3.703177213668823$ & $2.982294797897339$ \\
			\midrule[1.2pt]
			PLDRv51-110M-4 & $\checkmark$ & $\checkmark$ & $1.320483189459765 \times 10^{-8}$ & $2.601552391052246 \times 10^{1}$ & $2.768517255783081$ & $2.369031190872192$ \\
			\midrule
			PLDRv51-110M-4 & $\times$ & $\times$ & $1.311965114325631 \times 10^{-8}$ & $2.601552391052246 \times 10^{1}$ & $2.768517255783081$ & $2.369031190872192$ \\
			\midrule
			PLDRv51Gi-106M-4 & $\checkmark$ & NA & $1.300390373160099 \times 10^{-8}$ & $2.601552391052246 \times 10^{1}$ & $2.768517255783081$ & $2.369030952453613$ \\
			\midrule[1.2pt]
			PLDRv51-110M-5 & $\checkmark$ & $\checkmark$ & $8.018530479603214 \times 10^{-7}$ & $4.213090613484383 \times 10^{-2}$ & $2.386214017868042$ & $2.134617090225220$ \\
			\midrule
			PLDRv51-110M-5 & $\times$ & $\times$ & $8.018530479603214 \times 10^{-7}$ & $4.213090613484383 \times 10^{-2}$ & $2.386214017868042$ & $2.134617090225220$ \\
			\midrule
			PLDRv51Gi-106M-5 & $\checkmark$ & NA & $7.324443345169129 \times 10^{-7}$ & $4.213090613484383 \times 10^{-2}$ & $2.386213779449463$ & $2.134617090225220$ \\
			\midrule[1.2pt]
			PLDRv51-DAG-110M & $\checkmark$ & $\checkmark$ & $1.322519710811321 \times 10^{-5}$ & $7.740693981759250 \times 10^{-4}$ & $6.438840031623840  \times 10^{-1}$ & $5.846958756446838  \times 10^{-1}$ \\
			\midrule
			PLDRv51-DAG-110M & $\times$ & $\times$ & $1.206233173434157 \times 10^{-5}$ & $7.740693399682641 \times 10^{-4}$ & $6.438845396041870  \times 10^{-1}$ & $5.846965312957764  \times 10^{-1}$ \\
			\midrule
			PLDRv51Gi-DAG-106M & $\checkmark$ & NA & $1.065814103640150 \times 10^{-5}$ & $7.740692235529423  \times 10^{-4}$ & $6.438848972320557 \times 10^{-1}$ & $5.846968889236450 \times 10^{-1}$ \\
			\bottomrule[1.2pt]
		\end{tabular}
	}
\end{table}

\begin{table}[!htb]
	\centering
	\caption{Maximum of absolute value of determinants of heads from deductive outputs of PLDR-LLM among all layers. Deductive outputs were inferred as described in table \ref{table2}. $\nearrow$ indicates value is beyond the maximum value for the floating-point variable (overflow). Values are shown up to 15 decimal digits.}
	\label{table3}
	\resizebox{0.8\textwidth}{!}{
		\begin{tabular}{c c c c c c c}
			\toprule[1.2pt]
			Model & \multicolumn{2}{c}{Cache} & \multicolumn{4}{c}{Maximum Magnitude of Determinant} \\
			\cmidrule(lr){2-3} \cmidrule(lr){4-7}
			 & KV & G & $\tA$ & $\tA_{LM}$ & $\tA_{\textbf{P}}$ & $\tG_{LM}$ \\
			\midrule[1.2pt]
			PLDRv51-104M & $\checkmark$ & $\checkmark$ & 0 & 0 & $\nearrow$ & $\nearrow$ \\
			\midrule
			PLDRv51-104M & $\times$ & $\times$ & 0 & 0 & $\nearrow$ & $\nearrow$ \\
			\midrule
 			PLDRv51Gi-99M & $\checkmark$ & NA & 0 & 0 & $\nearrow$ & $\nearrow$ \\
			\midrule[1.2pt]
			PLDRv51-110M-1 & $\checkmark$ & $\checkmark$ & 0 & 0 & $\nearrow$ & $8.379209065368125 \times 10^{17}$ \\
			\midrule
			PLDRv51-110M-1 & $\times$ & $\times$ & 0 & 0 & $\nearrow$ & $8.379209065368125 \times 10^{17}$ \\
			\midrule
			PLDRv51Gi-106M-1 & $\checkmark$ & NA & 0 & 0 & $\nearrow$ & $8.379209065368125 \times 10^{17}$ \\
			\midrule[1.2pt]
			PLDRv51-110M-2 & $\checkmark$ & $\checkmark$ & 0 & 0 & $\nearrow$ & $8.224172600092262  \times 10^{16}$ \\
			\midrule
			PLDRv51-110M-2 & $\times$ & $\times$ & 0 & 0 & $\nearrow$ & $8.224172600092262  \times 10^{16}$ \\
			\midrule
			PLDRv51Gi-106M-2 & $\checkmark$ & NA & 0 & 0 & $\nearrow$ & $8.224172600092262 \times 10^{16}$ \\
			\midrule[1.2pt]
			PLDRv51-110M-3 & $\checkmark$ & $\checkmark$ & 0 & 0 & $\nearrow$ & $6.558759685657310  \times 10^{25}$ \\
			\midrule
			PLDRv51-110M-3 & $\times$ & $\times$ & 0 & 0 & $\nearrow$ & $6.558759685657310  \times 10^{25}$ \\
			\midrule
			PLDRv51Gi-106M-3 & $\checkmark$ & NA & 0 & 0 & $\nearrow$ & $6.558759685657310 \times 10^{25}$ \\
			\midrule[1.2pt]
			PLDRv51-110M-4 & $\checkmark$ & $\checkmark$ & 0 & 0 & $\nearrow$ & $6.680977184587632  \times 10^{31}$ \\
			\midrule
			PLDRv51-110M-4 & $\times$ & $\times$ & 0 & 0 & $\nearrow$ & $6.680999912393041  \times 10^{31}$ \\
			\midrule
			PLDRv51Gi-106M-4 & $\checkmark$ & NA & 0 & 0 & $\nearrow$ & $6.680975250306321 \times 10^{31}$ \\
			\midrule[1.2pt]
			PLDRv51-110M-5 & $\checkmark$ & $\checkmark$ & 0 & 0 & $\nearrow$ & $\nearrow$ \\
			\midrule
			PLDRv51-110M-5 & $\times$ & $\times$ & 0 & 0 & $\nearrow$ & $\nearrow$ \\
			\midrule
			PLDRv51Gi-106M-5 & $\checkmark$ & NA & 0 & 0 & $\nearrow$ & $\nearrow$ \\
			\midrule[1.2pt]
			PLDRv51-DAG-110M & $\checkmark$ & $\checkmark$ & 0 & 0 & $2.462903110682549 \times 10^{-34}$ & $2.638091745055249 \times 10^{-9}$ \\
			\midrule
			PLDRv51-DAG-110M & $\times$ & $\times$ & 0 & 0 & $2.464753825518873 \times 10^{-34}$ & $2.638115947917186 \times 10^{-9}$ \\
			\midrule
			PLDRv51Gi-DAG-106M & $\checkmark$ & NA &0 & 0 & $2.461961567258190 \times 10^{-34}$ & $2.638097518214977 \times 10^{-9}$ \\
			\bottomrule[1.2pt]
		\end{tabular}
	}
\end{table}

\subsection{Benchmark Evaluation}

The zero-shot performance of PLDR-LLMs with different model parameters is shown in table \ref{table4} with different KV-cache and G-cache configurations. The benchmark scores are same for all datasets with or without caching, as a consequence of highly invariant, generalizable deductive output characteristics of PLDR-LLM. The benchmark scores are comparable to the reference models reported in literature. The negative test model evaluations show reduced scores on average, indicating that the $\tG_{LM}$ has an actual effect on improving benchmark scores. 

\begin{table}[!htb]
	\caption{Zero-shot benchmark scores with different cache settings and negative test score results with mismatched $\tG_{LM}$ values. Benchmark scores evaluated on reference LLMs of similar parameter size are also shown. HS: Hellaswag, OBQA: OpenBookQA, WG: WinoGrande, TQA: TruthfulQA.}
	\label{table4}
	\centering
	\resizebox{\textwidth}{!}{
		\begin{tabular}{c c c c c c c c c c c c c}
			\toprule[1.2pt]
			Model & \multicolumn{2}{c}{Cache} & \multicolumn{10}{c}{Benchmark Score} \\
			\cmidrule(lr){2-3} \cmidrule(lr){4-13} 
			& KV & G & ARC-c & ARC-e & HS & OBQA & PIQA & SIQA & WG & Avg-1 & TQA & Avg-2 \\
			\midrule[1.2pt]
			PLDRv51-104M & $\checkmark$ & $\checkmark$ & 22.95 & 37.58 & 29.07 & 25.00 & 61.92 & 42.02 & 50.83 & 38.48 & 45.59 & 39.37 \\
			\midrule
			PLDRv51-104M & $\times$ & $\times$ & 22.95 & 37.58 & 29.07 & 25.00 & 61.92 & 42.02 & 50.83 & 38.48 & 45.59 & 39.37 \\
			\midrule
			PLDRv51Gi-99M & $\checkmark$ & NA & 22.95 & 37.58 & 29.07 & 25.00 & 61.92 & 42.02 & 50.83 & 38.48 & 45.59 & 39.37 \\
			\midrule[1.2pt]
			PLDRv51-110M-1 & $\checkmark$ & $\checkmark$ & 21.25 & 36.28 & 29.23 & 26.60 & 61.86 & 42.12 & 49.88 & 38.17 & 45.88 & 39.14 \\
			\midrule
			PLDRv51-110M-1 & $\times$ & $\times$ & 21.25 & 36.28 & 29.23 & 26.60 & 61.86 & 42.12 & 49.88 & 38.17 & 45.88 & 39.14 \\
			\midrule
			PLDRv51Gi-106M-1 & $\checkmark$ & NA & 21.25 & 36.28 & 29.23 & 26.60 & 61.86 & 42.12 & 49.88 & 38.17 & 45.88 & 39.14 \\
			\midrule[1.2pt]
			PLDRv51-DAG-110M & $\checkmark$ & $\checkmark$ & 21.93 & 35.44 & 28.69 & 26.60 & 61.04 & 41.45 & 51.62 & 38.11 & 45.92 & 39.09 \\
			\midrule
			PLDRv51-DAG-110M & $\times$ & $\times$ & 21.93 & 35.44 & 28.69 & 26.60 & 61.04 & 41.45 & 51.62 & 38.11 & 45.92 & 39.09 \\
			\midrule
			PLDRv51Gi-DAG-106M & $\checkmark$ & NA & 21.93 & 35.44 & 28.69 & 26.60 & 61.04 & 41.45 & 51.62 & 38.11 & 45.92 & 39.09 \\
			\midrule[1.2pt]
			GPT2-124M & $\checkmark$ & NA & 22.70 & 39.48 & 31.14 & 27.20 & 62.51 & 41.15 & 50.59 & 39.25 & 40.69 & 39.43 \\
			\midrule
			GPT-Neo-125M & $\checkmark$ & NA & 23.12 & 39.39 & 30.40 & 26.20 & 62.46 & 42.07 & 50.91 & 39.22 & 45.58 & 40.02 \\
			\midrule
			Phytia-160M & $\checkmark$ & NA & 24.66 & 38.47 & 31.22 & 27.60 & 61.64 & 40.69 & 51.07 & 39.33 & 44.15 & 39.94 \\
			\midrule[1.2pt]
			PLDRv51-106M-1-NAB1 & $\checkmark$ & $\times$ & 22.35 & 32.20 & 26.56 & 26.80 & 56.37 & 39.76 & 48.86 & 36.13 & 49.21 & 37.76 \\
			\midrule
			PLDRv51-106M-1-NAB2 & $\checkmark$ & $\times$ & 22.53 & 32.58 & 26.47 & 25.60 & 55.98 & 40.12 & 48.07 & 35.91 & 49.16 & 37.56 \\
			\bottomrule[1.2pt]
		\end{tabular}
	}
\end{table}

Table \ref{table5} shows evaluation of benchmarks with zero-shot setting for the models with higher SwiGLU:LU ratios and their learning rates/warm up step sizes adjusted for each model. All models show identical scores with and without caches available. PLDRv51-110M-3 has the smallest learning rate/warm up step size configuration among all PLDRv51 models with same layer size and number of attention heads, and shows the highest Avg-1 and lowest Avg-2 scores among these models, particularly impacted by a low TruthfulQA score.

We compared the inference time of PLDRv51-104M and PLDRv51-110M-1 with reference model GPT-Neo-125M with KV-cache enabled in table \ref{table6}. The KV-cache and G-cache improve the inference time by a factor of 3 when enabled for PLDR-LLMs. With KV-cache and G-cache enabled, PLDRv51-104M has up to 27\%  and PLDRv51-110M-1 has up to 39\% faster inference time compared to GPT-Neo-125M.

\begin{table}[!htb]
	\caption{Zero-shot Benchmark evaluation results at different cache settings for PLDR-LLMs with increased SwiGLU:LU ratios. HS: Hellaswag, OBQA: OpenBookQA, WG: WinoGrande, TQA: TruthfulQA.}
	\label{table5}
	\centering
	\resizebox{\textwidth}{!}{
		\begin{tabular}{c c c c c c c c c c c c c}
			\toprule[1.2pt]
			Model & \multicolumn{2}{c}{Cache} & \multicolumn{10}{c}{Benchmark Score} \\
			\cmidrule(lr){2-3} \cmidrule(lr){4-13} 
			& KV & G & ARC-c & ARC-e & HS & OBQA & PIQA & SIQA & WG & Avg-1 & TQA & Avg-2 \\
			\midrule[1.2pt]
			PLDRv51-110M-3  & $\checkmark$ & $\checkmark$ & 21.33 & 36.20 & 29.43 & 27.60 & 62.73 & 41.71 & 49.96 & 38.42 & 42.62 & 38.95 \\
			\midrule
			PLDRv51-110M-3 & $\times$ & $\times$ & 21.33 & 36.20 & 29.43 & 27.60 & 62.73 & 41.71 & 49.96 & 38.42 & 42.62 & 38.95 \\
			\midrule
			PLDRv51Gi-106M-3 & $\checkmark$ & NA &21.33 & 36.20 & 29.43 & 27.60 & 62.73 & 41.71 & 49.96 & 38.42 & 42.62 & 38.95 \\
			\midrule[1.2pt]
			PLDRv51-110M-4  & $\checkmark$ & $\checkmark$ & 21.93 & 35.77 & 28.98 & 27.20 & 62.24 & 42.37 & 49.64 & 38.31 & 44.37 & 39.06 \\
			\midrule
			PLDRv51-110M-4 & $\times$ & $\times$ & 21.93 & 35.77 & 28.98 & 27.20 & 62.24 & 42.37 & 49.64 & 38.31 & 44.37 & 39.06 \\
			\midrule
			PLDRv51Gi-106M-4 & $\checkmark$ & NA &21.93 & 35.77 & 28.98 & 27.20 & 62.24 & 42.37 & 49.64 & 38.31 & 44.37 & 39.06 \\
			\midrule[1.2pt]	
			PLDRv51-110M-5 & $\checkmark$ & $\checkmark$ & 21.50 & 36.36 & 28.80 & 26.40 & 61.70 & 42.17 & 49.80 & 38.10 & 45.54 & 39.03 \\
			\midrule
			PLDRv51-110M-5 & $\times$ & $\times$ & 21.50 & 36.36 & 28.80 & 26.40 & 61.70 & 42.17 & 49.80 & 38.10 & 45.54 & 39.03 \\
			\midrule
			PLDRv51Gi-106M-5 & $\checkmark$ & NA &21.50 & 36.36 & 28.80 & 26.40 & 61.70 & 42.17 & 49.80 & 38.10 & 45.54 & 39.03 \\
			\bottomrule[1.2pt]
		\end{tabular}
	}
\end{table}

\begin{table}[!htb]
	\caption{Inference time of PLDR-LLMs with and without caching, compared to a reference model of similar parameter size. The prompt was "Write a letter requesting people use language models responsibly." and the continuation was generated for 100 tokens with nucleus sampling (top-p=0.8) for PLDR-LLMs and a reference model from the literature. 10 runs of 100 generation loops was performed in each case. More details can be found in the appendix.}
	\label{table6}
	\centering
	\resizebox{0.6\textwidth}{!}{
		\begin{tabular}{c c c c}
			\toprule[1.2pt]
			Model & KV-cache & G-cache & Inference Time (ms) \\
			\midrule[1.2pt]
			PLDRv51-104M & $\times$ & $\times$ & $1700 \pm 82$ \\
			\midrule
			PLDRv51-104M & $\checkmark$ & $\checkmark$ & $579 \pm 14.8$ \\
			\midrule
			PLDRv51-110M-1 & $\times$ & $\times$ & $1390 \pm 60.8$ \\
			\midrule
			PLDRv51-110M-1 & $\checkmark$ & $\checkmark$ & $484 \pm 7.52$ \\
			\midrule
			GPT-Neo-125M & $\checkmark$ & NA & $788 \pm 7.39$ \\
			\bottomrule[1.2pt]
		\end{tabular}
	}
\end{table}

The evaluation of PLDR-LLMs with predefined $\tG_{LM}$ are shown on table \ref{table7}, together with PLDRv51-110M-2 which has a trainable $\tG_{LM}$ and pretrained on the same data interval. PLDRv51-110M-2 has the highest average score with a slight lead, followed by PLDRv51G-106M-1 with transferred $\tG_{LM}$, PLDRv51G-106M-2 equivalent to an LLM with SDPA and PLDRv51G-106M-3 with randomly distributed $\tG_{LM}$, respectively.

\begin{table}[!htb]
	\caption{Zero-shot benchmark scored for learnable and pre-defined $\tG_{LM}$ tensor on the RefinedWeb data interval [16M, 32M]. All models were evaluated with KV-cache and G-cache enabled when available. HS: Hellaswag, OBQA: OpenBookQA, WG: WinoGrande, TQA: TruthfulQA.}
	\label{table7}
	\centering
	\resizebox{\textwidth}{!}{
		\begin{tabular}{c c c c c c c c c c c}
			\toprule[1.2pt]
			Model & \multicolumn{10}{c}{Benchmark Score} \\
			\cmidrule(lr){2-11} 
			& ARC-c & ARC-e & HS & OBQA & PIQA & SIQA & WG & Avg-1 & TQA & Avg-2 \\
			\midrule[1.2pt]
			PLDRv51-110M-2 & 21.76 & 37.21 & 29.32 & 26.80 & 62.19 & 41.30 & 49.96 & 38.36 & 45.45 & 39.25 \\
			\midrule
			PLDRv51G-106M-1 & 21.76 & 36.91 & 28.43 & 25.00 & 62.08 & 42.22 & 51.22 & 38.23 & 45.59 & 39.15 \\
			\midrule
			PLDRv51G-106M-2 & 22.18 & 35.73 & 28.77 & 25.20 & 62.13 & 42.32 & 51.46 & 38.26 & 44.72 & 39.06 \\
			\midrule
			PLDRv51G-106M-3 & 21.76 & 36.20 & 28.71 & 26.40 & 61.15 & 40.69 & 50.51 & 37.92 & 45.78 & 38.90 \\
			\bottomrule[1.2pt]
		\end{tabular}
	}
\end{table}

\begin{figure}[!htb]
	\centering
	\begin{subfigure}[b]{0.45\textwidth}
		\centering
		\includegraphics[width=1\textwidth]{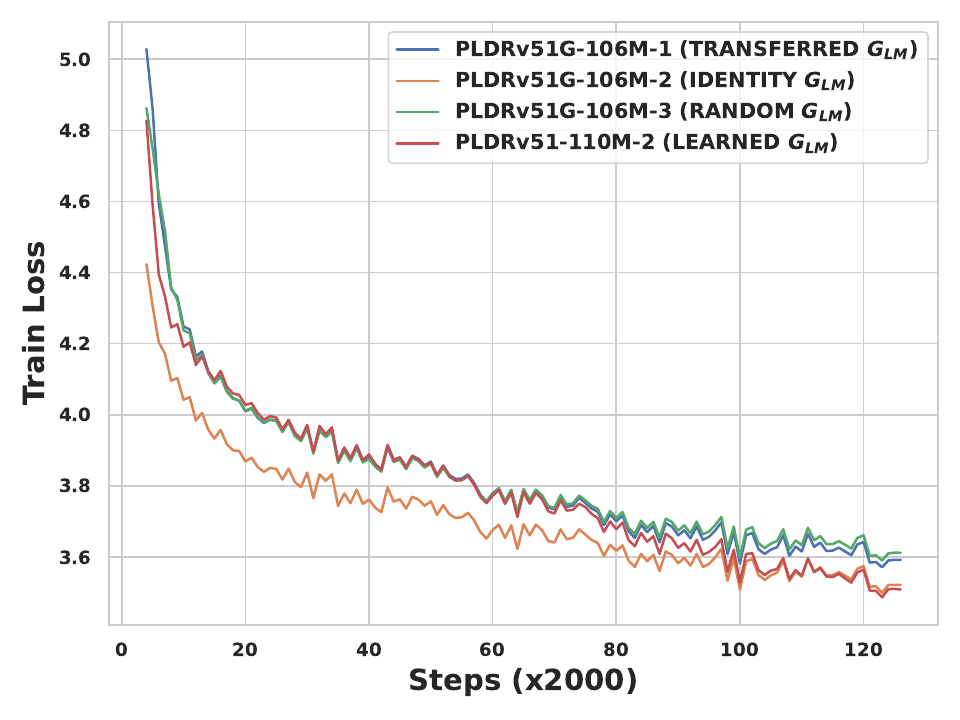}
		\caption{}
		\label{fig1a}
	\end{subfigure}
	\hfill
	\begin{subfigure}[b]{0.45\textwidth}
		\centering
		\includegraphics[width=1\textwidth]{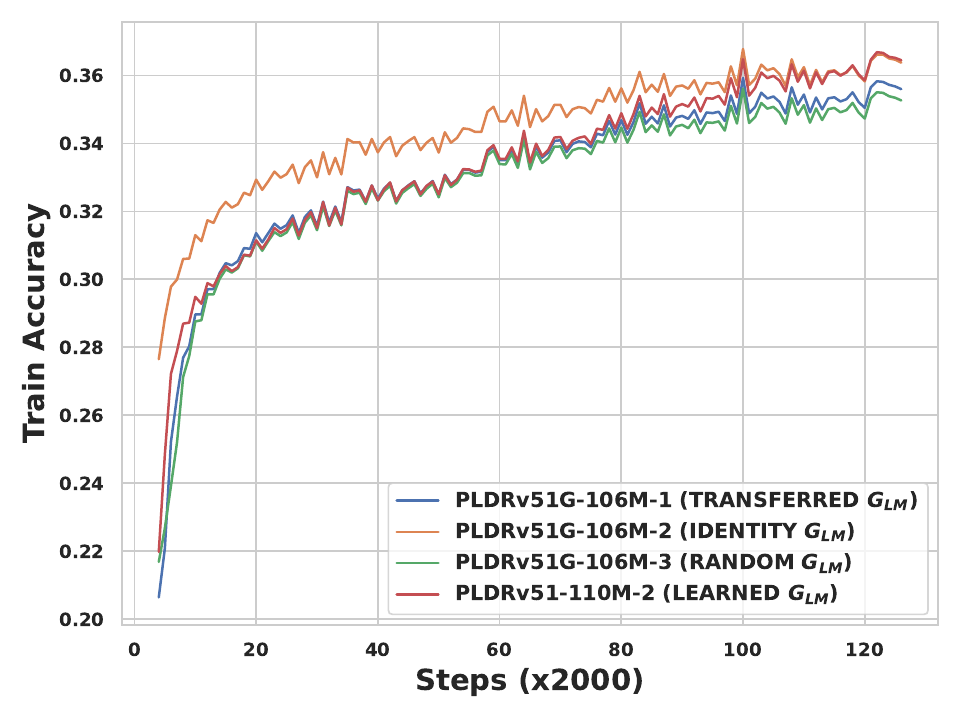}
		\caption{}
		\label{fig1b}
	\end{subfigure}
	\hfill
	\begin{subfigure}[b]{0.45\textwidth}
		\centering
		\includegraphics[width=1\textwidth]{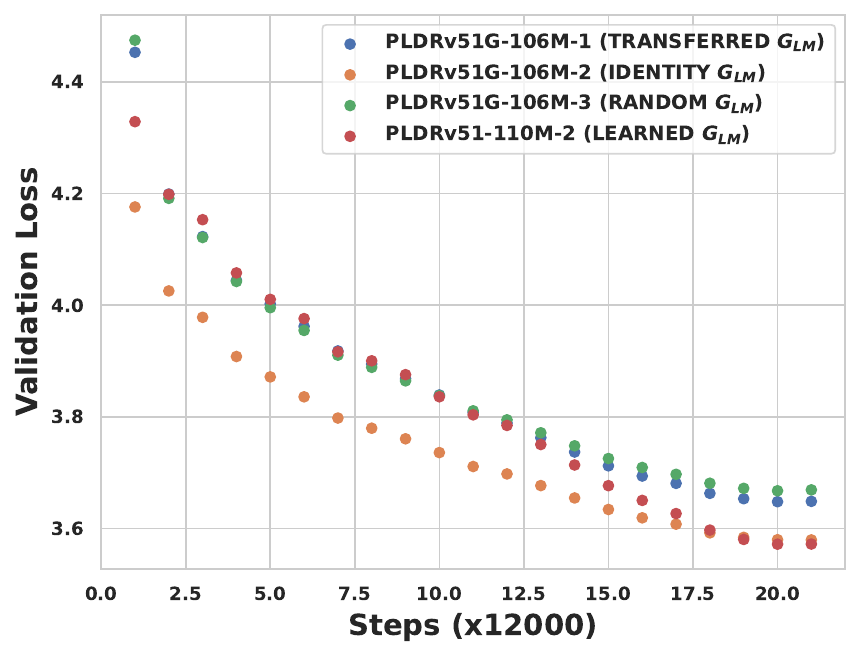}
		\caption{}
		\label{fig1c}
	\end{subfigure}
	\hfill
	\begin{subfigure}[b]{0.45\textwidth}
		\centering
		\includegraphics[width=1\textwidth]{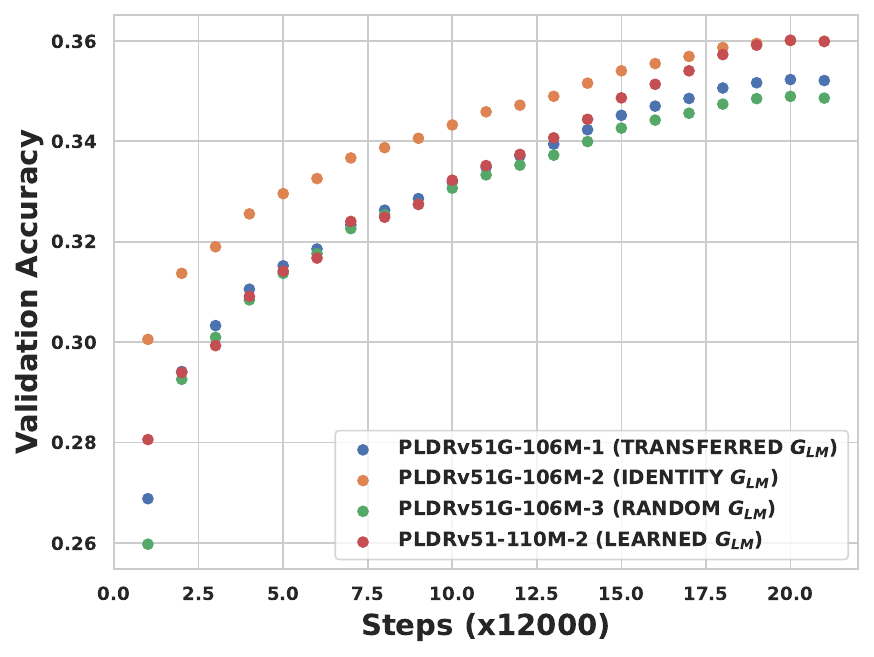}
		\caption{}
		\label{fig1d}
	\end{subfigure}
	\caption{Train and validation loss/accuracy curves for PLDR-LLMs in table \ref{table7}. Train loss is captured as a running loss at every 2000 steps. Validation loss is measured at every 12000 steps using 2000 batches/rank from part of RefinedWeb dataset that is not used in pretraining.}
	\label{fig1}
\end{figure}

The loss and accuracy curves of the models with benchmark scores on table \ref{table7} are shown in figure \ref{fig1}. The PLDRv51G-106M-1 and PLDRv51G-106M-3 follow very similar loss and accuracy values. PLDRv51G-106M-2 follows typically a lower loss and higher accuracy curve. Compared to the models with predefined $\tG_{LM}$, PLDRv51-110M-2 with learnable $\tG_{LM}$ during training exhibits a unique characteristic such that it follows similar loss and accuracy trend as PLDRv51G-106M-1 for the first $\sim$180k steps after which it gradually aligns with PLDRv51G-106M-2 for these curves. This indicates that the $\tG_{LM}$ learned through deep PLGA net during training is distinct from predefined and constant $\tG_{LM}$. The model learns a $\tG_{LM}$ tensor unique to the dataset it was pretrained on. It also emphasizes that the PLDR-LLM is a foundational model of which LLM with SDPA is a special case where $\tG_{LM}$ is set as identity.

\section{Discussion}

The deductive outputs were introduced in \citep{Gokden2021} to demonstrate that local and global characteristics of the representation space can be observed and investigated through them. Out of these deductive outputs, $\tG_{LM}$ ($\tA_{LM}$)  were output of a network composed of deep residual and fully-connected layers and represent localized characteristics of a sample from the dataset as it is required to infer them. On the other hand, learnable parameters such as the power coefficients, and custom weight/bias parameters represent the generalizable (and global) characteristics learned from the entire dataset. Empirical observations we presented here indicating that $\tG_{LM}$ ($\tA_{LM}$) is in fact a generalizable representation of the dataset have important implications. If a single input sample is considered as a local variable, $\tG_{LM}$ is also a local but invariant variable of that local frame up to a small perturbation and differs from learned model weights in this aspect. PLGA was inspired at the intersection of quantization of samples through tokenization and their dense features in an N-dimensional embedding space, collectively representing the nodes and their feature vectors as part of a graph. The interactions between graph nodes are determined by potentials that shape the high-dimensional loss landscape similar to how mass and energy curves space-time. The effect we see here appears to be an empirical manifestation  of Mach's principle \citep{Misner1973Gravitation} in the embedding space and it can be stated as "local inertial frames are affected by the distribution of matter and energy everywhere". $\tG_{LM}$ is learned and modified non-linearly by all the samples that model is pretrained on, yet it is invariant to a high degree and a generalizable linear operator for the local frame of any input sample at inference. In other words, the input variable generates $\tG_{LM}$ locally, but because it is invariant to all inputs up to a small perturbation, it is almost indistinguishable from a constant linear operator and can be cached as such.

The observation that we can replace a portion of neural network with a generalizable tensor operator means that there is a fundamental asymmetry between training and inference phases. Although the evaluation is identical during inference after this replacement up to a small perturbation, it would not be possible to train a model as effective without a fully defined PLGA net as it was shown in results on table \ref{table7} and figure \ref{fig1}. A practical application of this would be to conceal the PLGA weights during inference.

\section{Conclusion}

PLDR-LLM is a new type of foundational model that can learn a generalizable deductive output as an invariant tensor operator up to a small perturbation and this operator can replace its generating neural network at inference. This characteristic also makes it possible to use KV-cache optimizations more efficiently for faster inference by switching to an invariant $\tG_{LM}$ as tensor operator after inferring it with initial prompt only once. We showed that this observation holds for a very high degree of fidelity after caching for deductive outputs and for a variety of benchmarks with zero-shot setting. We also compared the model performance by transfer learning an already inferred $\tG_{LM}$, and with predefined $\tG_{LM}$ equal to identity and a random tensor from a normal distribution with unit variance and zero mean. The benchmark results and loss and accuracy curves show that PLDR-LLM with its full PLGA network is distinct and better performing than $\tG_{LM}$ with predefined values. The LLM with SDPA widely used in literature is a special case of PLDR-LLM where $\tG_{LM}$ is set to identity. PLDR-LLM exhibits an asymmetry between training and inference phases through caching that is unique to this foundational model architecture. Common characteristics of deductive outputs across models pretrained in this paper indicate that the model learns a generalizable singularity condition for the deductive outputs that leads to this asymmetry.

\section*{Acknowledgments}

I am grateful to my parents for their support and patience. This research was conducted independently without support from a grant or corporation.

\bibliographystyle{unsrtnat}
\bibliography{pldrllm-kvgcache-references}

\begin{thebibliography}{25}
\providecommand{\natexlab}[1]{#1}
\providecommand{\url}[1]{\texttt{#1}}
\expandafter\ifx\csname urlstyle\endcsname\relax
  \providecommand{\doi}[1]{doi: #1}\else
  \providecommand{\doi}{doi: \begingroup \urlstyle{rm}\Url}\fi

\bibitem[Gokden(2024)]{Gokden2024pldrllm}
Burc Gokden.
\newblock Pldr-llm: Large language model from power law decoder
  representations, 2024.
\newblock URL \url{https://arxiv.org/abs/2410.16703}.

\bibitem[Gokden(2021)]{Gokden2021}
Burc Gokden.
\newblock Power law graph transformer for machine translation and
  representation learning, 2021.
\newblock URL \url{https://arxiv.org/abs/2107.02039}.

\bibitem[Gokden(2019)]{Gokden2019}
Burc Gokden.
\newblock Coulgat: An experiment on interpretability of graph attention
  networks, 2019.
\newblock URL \url{https://arxiv.org/abs/1912.08409}.

\bibitem[Radford et~al.(2019)Radford, Wu, Child, Luan, Amodei, and
  Sutskever]{Radford2019gpt2}
Alec Radford, Jeff Wu, Rewon Child, David Luan, Dario Amodei, and Ilya
  Sutskever.
\newblock Language models are unsupervised multitask learners.
\newblock 2019.
\newblock URL
  \url{https://cdn.openai.com/better-language-models/language_models_are_unsupervised_multitask_learners.pdf}.

\bibitem[Touvron et~al.(2023{\natexlab{a}})Touvron, Lavril, Izacard, Martinet,
  Lachaux, Lacroix, Rozière, Goyal, Hambro, Azhar, Rodriguez, Joulin, Grave,
  and Lample]{Touvron2023llama}
Hugo Touvron, Thibaut Lavril, Gautier Izacard, Xavier Martinet, Marie-Anne
  Lachaux, Timothée Lacroix, Baptiste Rozière, Naman Goyal, Eric Hambro,
  Faisal Azhar, Aurelien Rodriguez, Armand Joulin, Edouard Grave, and Guillaume
  Lample.
\newblock Llama: Open and efficient foundation language models.
\newblock 2023{\natexlab{a}}.
\newblock URL \url{https://arxiv.org/abs/2302.13971}.

\bibitem[Touvron et~al.(2023{\natexlab{b}})Touvron, Martin, Stone, Albert,
  Almahairi, Babaei, Bashlykov, Batra, Bhargava, Bhosale, Bikel, Blecher,
  Ferrer, Chen, Cucurull, Esiobu, Fernandes, Fu, Fu, Fuller, Gao, Goswami,
  Goyal, Hartshorn, Hosseini, Hou, Inan, Kardas, Kerkez, Khabsa, Kloumann,
  Korenev, Koura, Lachaux, Lavril, Lee, Liskovich, Lu, Mao, Martinet, Mihaylov,
  Mishra, Molybog, Nie, Poulton, Reizenstein, Rungta, Saladi, Schelten, Silva,
  Smith, Subramanian, Tan, Tang, Taylor, Williams, Kuan, Xu, Yan, Zarov, Zhang,
  Fan, Kambadur, Narang, Rodriguez, Stojnic, Edunov, and
  Scialom]{Touvron2023llama2}
Hugo Touvron, Louis Martin, Kevin Stone, Peter Albert, Amjad Almahairi, Yasmine
  Babaei, Nikolay Bashlykov, Soumya Batra, Prajjwal Bhargava, Shruti Bhosale,
  Dan Bikel, Lukas Blecher, Cristian~Canton Ferrer, Moya Chen, Guillem
  Cucurull, David Esiobu, Jude Fernandes, Jeremy Fu, Wenyin Fu, Brian Fuller,
  Cynthia Gao, Vedanuj Goswami, Naman Goyal, Anthony Hartshorn, Saghar
  Hosseini, Rui Hou, Hakan Inan, Marcin Kardas, Viktor Kerkez, Madian Khabsa,
  Isabel Kloumann, Artem Korenev, Punit~Singh Koura, Marie-Anne Lachaux,
  Thibaut Lavril, Jenya Lee, Diana Liskovich, Yinghai Lu, Yuning Mao, Xavier
  Martinet, Todor Mihaylov, Pushkar Mishra, Igor Molybog, Yixin Nie, Andrew
  Poulton, Jeremy Reizenstein, Rashi Rungta, Kalyan Saladi, Alan Schelten, Ruan
  Silva, Eric~Michael Smith, Ranjan Subramanian, Xiaoqing~Ellen Tan, Binh Tang,
  Ross Taylor, Adina Williams, Jian~Xiang Kuan, Puxin Xu, Zheng Yan, Iliyan
  Zarov, Yuchen Zhang, Angela Fan, Melanie Kambadur, Sharan Narang, Aurelien
  Rodriguez, Robert Stojnic, Sergey Edunov, and Thomas Scialom.
\newblock Llama 2: Open foundation and fine-tuned chat models.
\newblock 2023{\natexlab{b}}.
\newblock URL \url{https://arxiv.org/abs/2307.09288}.

\bibitem[Shazeer(2019)]{Shazeer2019onewrite}
Noam Shazeer.
\newblock Fast transformer decoding: One write-head is all you need, 2019.
\newblock URL \url{https://arxiv.org/abs/1911.02150}.

\bibitem[Liu et~al.(2024)Liu, Yuan, Jin, Zhong, Xu, Braverman, Chen, and
  Hu]{Liu2024kivi}
Zirui Liu, Jiayi Yuan, Hongye Jin, Shaochen~(Henry) Zhong, Zhaozhuo Xu,
  Vladimir Braverman, Beidi Chen, and Xia Hu.
\newblock Kivi: a tuning-free asymmetric 2bit quantization for kv cache.
\newblock In \emph{Proceedings of the 41st International Conference on Machine
  Learning}, ICML'24, 2024.

\bibitem[Zhao et~al.(2023)Zhao, Gu, Varma, Luo, Huang, Xu, Wright, Shojanazeri,
  Ott, Shleifer, Desmaison, Balioglu, Damania, Nguyen, Chauhan, Hao, Mathews,
  and Li]{Zhao2023fsdp}
Yanli Zhao, Andrew Gu, Rohan Varma, Liang Luo, Chien-Chin Huang, Min Xu, Less
  Wright, Hamid Shojanazeri, Myle Ott, Sam Shleifer, Alban Desmaison, Can
  Balioglu, Pritam Damania, Bernard Nguyen, Geeta Chauhan, Yuchen Hao, Ajit
  Mathews, and Shen Li.
\newblock Pytorch fsdp: Experiences on scaling fully sharded data parallel.
\newblock \emph{Proc. VLDB Endow.}, 16\penalty0 (12):\penalty0 3848–3860,
  2023.
\newblock ISSN 2150-8097.
\newblock \doi{10.14778/3611540.3611569}.

\bibitem[Penedo et~al.(2023)Penedo, Malartic, Hesslow, Cojocaru, Alobeidli,
  Cappelli, Pannier, Almazrouei, and Launay]{Penedo2023falcon}
Guilherme Penedo, Quentin Malartic, Daniel Hesslow, Ruxandra Cojocaru, Hamza
  Alobeidli, Alessandro Cappelli, Baptiste Pannier, Ebtesam Almazrouei, and
  Julien Launay.
\newblock The refinedweb dataset for falcon llm: outperforming curated corpora
  with web data only.
\newblock In \emph{Proceedings of the 37th International Conference on Neural
  Information Processing Systems}, NIPS '23, Red Hook, NY, USA, 2023. Curran
  Associates Inc.

\bibitem[Kudo and Richardson(2018)]{Kudo2018sentencepiece}
Taku Kudo and John Richardson.
\newblock {S}entence{P}iece: A simple and language independent subword
  tokenizer and detokenizer for neural text processing.
\newblock In \emph{Proceedings of the 2018 Conference on Empirical Methods in
  Natural Language Processing: System Demonstrations}, pages 66--71, Brussels,
  Belgium, November 2018. Association for Computational Linguistics.
\newblock \doi{10.18653/v1/D18-2012}.
\newblock URL \url{https://aclanthology.org/D18-2012}.

\bibitem[Kudo(2018)]{Kudo2018subword}
Taku Kudo.
\newblock Subword regularization: Improving neural network translation models
  with multiple subword candidates.
\newblock In Iryna Gurevych and Yusuke Miyao, editors, \emph{Proceedings of the
  56th Annual Meeting of the Association for Computational Linguistics (Volume
  1: Long Papers)}, pages 66--75, Melbourne, Australia, July 2018. Association
  for Computational Linguistics.
\newblock \doi{10.18653/v1/P18-1007}.
\newblock URL \url{https://aclanthology.org/P18-1007}.

\bibitem[Black et~al.(2021)Black, Leo, Wang, Leahy, and Biderman]{gptneo}
Sid Black, Gao Leo, Phil Wang, Connor Leahy, and Stella Biderman.
\newblock {GPT-Neo: Large Scale Autoregressive Language Modeling with
  Mesh-Tensorflow}, March 2021.
\newblock URL \url{https://doi.org/10.5281/zenodo.5297715}.

\bibitem[Gao et~al.(2020)Gao, Biderman, Black, Golding, Hoppe, Foster, Phang,
  He, Thite, Nabeshima, Presser, and Leahy]{Gao2020pile}
Leo Gao, Stella Biderman, Sid Black, Laurence Golding, Travis Hoppe, Charles
  Foster, Jason Phang, Horace He, Anish Thite, Noa Nabeshima, Shawn Presser,
  and Connor Leahy.
\newblock The pile: An 800gb dataset of diverse text for language modeling.
\newblock 2020.
\newblock URL \url{https://arxiv.org/abs/2101.00027}.

\bibitem[Biderman et~al.(2023)Biderman, Schoelkopf, Anthony, Bradley, O'Brien,
  Hallahan, Khan, Purohit, Prashanth, Raff, Skowron, Sutawika, and Van
  Der~Wal]{Biderman2023pythia}
Stella Biderman, Hailey Schoelkopf, Quentin Anthony, Herbie Bradley, Kyle
  O'Brien, Eric Hallahan, Mohammad~Aflah Khan, Shivanshu Purohit, USVSN~Sai
  Prashanth, Edward Raff, Aviya Skowron, Lintang Sutawika, and Oskar Van
  Der~Wal.
\newblock Pythia: a suite for analyzing large language models across training
  and scaling.
\newblock In \emph{Proceedings of the 40th International Conference on Machine
  Learning}, ICML'23. JMLR.org, 2023.

\bibitem[Gao et~al.(2024)Gao, Tow, Abbasi, Biderman, Black, DiPofi, Foster,
  Golding, Hsu, Le~Noac'h, Li, McDonell, Muennighoff, Ociepa, Phang, Reynolds,
  Schoelkopf, Skowron, Sutawika, Tang, Thite, Wang, Wang, and Zou]{evalharness}
Leo Gao, Jonathan Tow, Baber Abbasi, Stella Biderman, Sid Black, Anthony
  DiPofi, Charles Foster, Laurence Golding, Jeffrey Hsu, Alain Le~Noac'h,
  Haonan Li, Kyle McDonell, Niklas Muennighoff, Chris Ociepa, Jason Phang,
  Laria Reynolds, Hailey Schoelkopf, Aviya Skowron, Lintang Sutawika, Eric
  Tang, Anish Thite, Ben Wang, Kevin Wang, and Andy Zou.
\newblock A framework for few-shot language model evaluation, 07 2024.
\newblock URL \url{https://zenodo.org/records/12608602}.

\bibitem[Clark et~al.(2018)Clark, Cowhey, Etzioni, Khot, Sabharwal, Schoenick,
  and Tafjord]{Clark2018arc}
Peter Clark, Isaac Cowhey, Oren Etzioni, Tushar Khot, Ashish Sabharwal, Carissa
  Schoenick, and Oyvind Tafjord.
\newblock Think you have solved question answering? try arc, the ai2 reasoning
  challenge.
\newblock 2018.
\newblock URL \url{https://arxiv.org/abs/1803.05457}.

\bibitem[Zellers et~al.(2019)Zellers, Holtzman, Bisk, Farhadi, and
  Choi]{Zellers2019hellaswag}
Rowan Zellers, Ari Holtzman, Yonatan Bisk, Ali Farhadi, and Yejin Choi.
\newblock Hellaswag: Can a machine really finish your sentence?
\newblock In \emph{Proceedings of the 57th Annual Meeting of the Association
  for Computational Linguistics}, 2019.

\bibitem[Sakaguchi et~al.(2021)Sakaguchi, Bras, Bhagavatula, and
  Choi]{Keisuke2019winogrande}
Keisuke Sakaguchi, Ronan~Le Bras, Chandra Bhagavatula, and Yejin Choi.
\newblock Winogrande: an adversarial winograd schema challenge at scale.
\newblock \emph{Commun. ACM}, 64\penalty0 (9):\penalty0 99–106, August 2021.
\newblock ISSN 0001-0782.
\newblock \doi{10.1145/3474381}.
\newblock URL \url{https://doi.org/10.1145/3474381}.

\bibitem[Lin et~al.(2022)Lin, Hilton, and Evans]{Lin2021truthfulqa}
Stephanie Lin, Jacob Hilton, and Owain Evans.
\newblock {T}ruthful{QA}: Measuring how models mimic human falsehoods.
\newblock In \emph{Proceedings of the 60th Annual Meeting of the Association
  for Computational Linguistics (Volume 1: Long Papers)}, pages 3214--3252,
  Dublin, Ireland, May 2022. Association for Computational Linguistics.
\newblock \doi{10.18653/v1/2022.acl-long.229}.
\newblock URL \url{https://aclanthology.org/2022.acl-long.229}.

\bibitem[Mihaylov et~al.(2018)Mihaylov, Clark, Khot, and
  Sabharwal]{Mihaylov2018obqa}
Todor Mihaylov, Peter Clark, Tushar Khot, and Ashish Sabharwal.
\newblock Can a suit of armor conduct electricity? a new dataset for open book
  question answering.
\newblock In \emph{EMNLP}, 2018.

\bibitem[Bisk et~al.(2020)Bisk, Zellers, Bras, Gao, and Choi]{Bisk2020piqa}
Yonatan Bisk, Rowan Zellers, Ronan~Le Bras, Jianfeng Gao, and Yejin Choi.
\newblock Piqa: Reasoning about physical commonsense in natural language.
\newblock In \emph{Thirty-Fourth AAAI Conference on Artificial Intelligence},
  2020.

\bibitem[Sap et~al.(2019)Sap, Rashkin, Chen, Le~Bras, and Choi]{Sap2019siqa}
Maarten Sap, Hannah Rashkin, Derek Chen, Ronan Le~Bras, and Yejin Choi.
\newblock Social {IQ}a: Commonsense reasoning about social interactions.
\newblock In \emph{Proceedings of the 2019 Conference on Empirical Methods in
  Natural Language Processing and the 9th International Joint Conference on
  Natural Language Processing (EMNLP-IJCNLP)}, pages 4463--4473, Hong Kong,
  China, November 2019. Association for Computational Linguistics.
\newblock \doi{10.18653/v1/D19-1454}.
\newblock URL \url{https://aclanthology.org/D19-1454}.

\bibitem[Misner et~al.(1973)Misner, Thorne, and Wheeler]{Misner1973Gravitation}
Charles~W. Misner, K.~S. Thorne, and J.~A. Wheeler.
\newblock \emph{{Gravitation}}, chapter 21.12, pages 543--551.
\newblock W. H. Freeman, San Francisco, 1973.
\newblock ISBN 978-0-691-17779-3.

\bibitem[Maas et~al.(2011)Maas, Daly, Pham, Huang, Ng, and Potts]{Maas2011imdb}
Andrew~L. Maas, Raymond~E. Daly, Peter~T. Pham, Dan Huang, Andrew~Y. Ng, and
  Christopher Potts.
\newblock Learning word vectors for sentiment analysis.
\newblock In \emph{Proceedings of the 49th Annual Meeting of the Association
  for Computational Linguistics: Human Language Technologies}, pages 142--150,
  Portland, Oregon, USA, June 2011. Association for Computational Linguistics.
\newblock URL \url{http://www.aclweb.org/anthology/P11-1015}.

\end{thebibliography}

\clearpage
\appendix

\section*{Appendix}
\section{Derivation of Number of Trainable Parameters for Power Law Graph Attention}

The ratio (\#ResL)/(\#A) in table \ref{table1} is the ratio of number of trainable parameters of deep residual network section of Power Law Graph Attention to the resulting tensor $\tA$ which has a $d_{k} \times d_{k}$ size per head. The residual network consist of 8 residual layers. Each residual layer has 2 SwiGLU (with layer size $Ad_{ff}$) and LU units and a LayerNorm layer which has $2 \times d_{k}$ trainable parameters in the implementation. The residual network is shared among all heads in a layer.

\begin{equation}
	\frac{\text{\# Parameters of Residual Network}}{\text{\# Parameters of $\tA$ per head}} = \frac{(((Ad_{ff} \times d_{k}+Ad_{ff})\times2+(Ad_{ff} \times d_{k}+d_{k})) \times 2+2 \times d_{k})\times8}{d_{k} \times d_{k}}
\end{equation}

The total number of trainable parameters for Power Law Graph Attention layer that is replaced with $\tG_{LM}$ depends on the number of heads $h$ (custom weights/biases and power coefficients are for each head) and number of decoder layers $L$:

\begin{equation}
	\left((((Ad_{ff} \times d_{k}+Ad_{ff})\times2+(Ad_{ff} \times d_{k}+d_{k})) \times 2+ 2 \times d_{k})\times 8+ 5 \times d_{k} \times d_{k} \times h+2 \times d_{k}\right)\times L
\end{equation}
\section{Code Snippets for Inference Time Comparison}
Below snippets were run on a Jupyter notebook.

PLDR-LLM without cache:
\begin{verbatim}
    sentence="Write a letter requesting people use language models responsibly."
	
    %%timeit -r 10 -n 100 
    text, _, _=e2e_obj.generate_text(sentence, 
                                     temperature=1.0, top_k=0, top_p=0.8, 
                                     enable_kvcache=False, enable_Gcache=False, 
                                     Gcachelst_init=None,
                                     max_length=100, save_att=None)
	
\end{verbatim}

PLDR-LLM with cache:
\begin{verbatim}
    sentence="Write a letter requesting people use language models responsibly."
	
    %%timeit -r 10 -n 100 
    text, _, _=e2e_obj.generate_text(sentence, 
                                     temperature=1.0, top_k=0, top_p=0.8, 
                                     enable_kvcache=True, enable_Gcache=True, 
                                     Gcachelst_init=None,
                                     max_length=100, save_att=None)
	
\end{verbatim}

Reference LLM with cache:
\begin{verbatim}
    from transformers import pipeline

    generator = pipeline('text-generation', model='EleutherAI/gpt-neo-125M')
    prompt = "Write a letter requesting people use language models responsibly."

    %%timeit -r 10 -n 100
    generator(prompt, max_new_tokens=100, do_sample=True, 
              temperature=1.0, top_k=0, top_p=0.8, use_cache=True)
\end{verbatim}

\section{Benchmark Datasets}

\textbf{ARC}. The AI2 Reasoning Challenge (ARC) dataset consists of multiple-choice grade school questions from $3^{rd}$ to $9^{th}$ grade. It consists of an easy set and a challenge set. The challenge set contains the questions answered incorrectly by both a retrieval based algorithm and a word co-occurrence algorithm \citep{Clark2018arc}.

\textbf{Hellaswag}. Harder Endings, Longer contexts, and Low-shot Activities for Situations With Adversarial Generations dataset is a commonsense natural language inference dataset that was prepared using adversarial filtering to create problems that are challenging to models, yet easy for humans \citep{Zellers2019hellaswag}.

\textbf{WinoGrande}. WinoGrande is a more challenging version of Winograd Schema Challenge that is a commonsense reasoning benchmark based on a set of pronoun resolution problems designed to be unsolvable for statistical models that rely on selectional preferences or word associations \citep{Keisuke2019winogrande}.

\textbf{TruthfulQA}. TruthfulQA is a benchmark that aims to measure truthfullness of a model. It consists of questions covering 38 categories such as health, law, finance and politics. The model should avoid imitating human contexts in pretraining dataset to perform well, since the questions are selected from the ones humans would answer incorrectly due to a false belief or misconception \citep{Lin2021truthfulqa}.

\textbf{OpenBookQA}. OpenBookQA is a question answering dataset that consists of about 6000 questions accompanied with scientific facts. To answer the questions correctly the model needs to combine with extra common knowledge beyond the facts included in the dataset \citep{Mihaylov2018obqa}.

\textbf{PIQA}. Physical Interaction:Question Answering dataset is a physical commonsense benchmark that aims to evaluate model performance for concepts that are traditionally only seen or experienced in the real world \citep{Bisk2020piqa}.

\textbf{SIQA}. Social Intelligence QA dataset is a social commonsense reasoning benchmark that aims to evaluate model performance for social situations. It consists of 38000 multiple-choice questions for probing emotional and social intelligence in a variety of everyday situations \citep{Sap2019siqa}.

\textbf{IMDB Review}. IMDB Review dataset is a collection of 50000 reviews with each movie having no more than 30 reviews. It was compiled for sentiment analysis and consists of an even number of highly polarized negative ($\leq4$ out of $10$) and positive ($\geq7$ out of $10$) reviews \citep{Maas2011imdb}.

\newpage
\section{Sample Text Outputs from PLDR-LLMs}
Input is several sentences from the beginning of a review sample from IMDB Review dataset \citep{Maas2011imdb} appended with the phrase "What I would like to say is". Continuation is the generated text output from PLDR-LLMs with nucleus sampling at $\text{top-p}=0.8$. The model generates text for 256 tokens or until it encounters an end of sentence ("[END]") token.

\begin{table}[!htb]
	\tiny
	\label{table8}
	\centering
	\resizebox{\textwidth}{!}{
	\begin{tabular}{| p{5cm} | p{5cm} | }
		\multicolumn{2}{c}{\textbf{PLDRv51-110M-1 with KV-cache and G-cache enabled}} \\
		\hline 
		\textbf{Input} & \textbf{Continuation} \\ 
		\hline
		I love sci-fi and am willing to put up with a lot. Sci-fi movies/TV are usually underfunded, under-appreciated and misunderstood. I tried to like this, I really did, but it is to good TV sci-fi as Babylon 5 is to Star Trek (the original). Silly prosthetics, cheap cardboard sets, stilted dialogues, CG that doesn't match the background, and painfully one-dimensional characters cannot be overcome with a 'sci-fi' setting. What I would like to say is 
		&
		that it has to do with the way we as a society take things for granted. I find that it's all one big set of rules and the other place of comfort and satisfaction to be in it. In reality, I'm really just living the character and the story is so far from that, and I think it's interesting how so many of the characters go from simple to very complex to so complex, and it also makes me think that in modern sci-fi it's just a very complex set of characters. I think that there's an entirely different message in there. The older ones aren't really meant to be in modern entertainment, but rather, like fantasy, to a point where it seems to go along so well with the space cinema, and then to how come it isn't real. I think they really have something in common with the contemporary perspective, they have such stagnation as 'so great' that they could only dream about the idea of running around in a quick-coming by falling in love. What I'm interested in is a unique set of topics in the core of the series that are pretty interesting. I don't think that even the best movies (such as ''The War. 2
		\\
		\hline
	\end{tabular} } 
\end{table}

\begin{table}[!htb]
	\tiny
	\label{table9}
	\centering
	\resizebox{\textwidth}{!}{
		\begin{tabular}{| p{5cm} | p{5cm} | }
			\multicolumn{2}{c}{\textbf{PLDRv51-110M-1 with KV-cache and G-cache disabled}} \\
			\hline 
			\textbf{Input} & \textbf{Continuation} \\ 
			\hline
			I love sci-fi and am willing to put up with a lot. Sci-fi movies/TV are usually underfunded, under-appreciated and misunderstood. I tried to like this, I really did, but it is to good TV sci-fi as Babylon 5 is to Star Trek (the original). Silly prosthetics, cheap cardboard sets, stilted dialogues, CG that doesn't match the background, and painfully one-dimensional characters cannot be overcome with a 'sci-fi' setting. What I would like to say is 
			&
			that I think the production designer (the producers) is a creative person and he will go to a point where it is inevitable that he will use the computer to make it work, or he will probably work to re-purpose the characters, but he just can't see that the other actors are going to be trying to make them work, or he can't, or he can't. The purpose of the role is to encourage the 'skeptics' and encourage them to see how they can help the directors and writers put the idea into practice. However, the producers and writers, who are generally going to put up with the production, can't always pull together a creative team and always have to be willing to work. So far, the producers have not shied away from doing the characters they're trying to achieve with the story they're trying to do. The writers and producers have failed to take credit for what they've done. They've tried to get the dialogue across, and it's not worth the effort. With that said, the producers are going to have to work on this project as a team and not just as a business. When I would be writing the scripts, I would have to consider how much
			\\
			\hline
	\end{tabular} } 
\end{table}

\begin{table}[!htb]
	\tiny
	\label{table10}
	\centering
	\resizebox{\textwidth}{!}{
		\begin{tabular}{| p{5cm} | p{5cm} | }
			\multicolumn{2}{c}{\textbf{PLDRv51G-106M-1 with KV-cache enabled}} \\
			\hline 
			\textbf{Input} & \textbf{Continuation} \\ 
			\hline
			I love sci-fi and am willing to put up with a lot. Sci-fi movies/TV are usually underfunded, under-appreciated and misunderstood. I tried to like this, I really did, but it is to good TV sci-fi as Babylon 5 is to Star Trek (the original). Silly prosthetics, cheap cardboard sets, stilted dialogues, CG that doesn't match the background, and painfully one-dimensional characters cannot be overcome with a 'sci-fi' setting. What I would like to say is 
			&
			that you do not think that your plot is falling apart because there is a depth. The makers of Sci-fi have a lot of special talent to work with. There is a special prize, but I think that's a bonus. The set of 'types' of Sci-fi is about fun, and you can enjoy all of them on the one night you're on the other. The set was supposed to be a bit older than my previous cast. It's about a character who is basically a rock star and has more time with it than I would like. I'm not a fan of the sound effects though. While I was taking the stand, I found that it would be a great hit. It also did not suit me, because I don't think the result is satisfying. I'm definitely a fan of Sci-fi. In fact, it's just a series that I think is less irritating.
			\\
			\hline
	\end{tabular} } 
\end{table}

\begin{table}[!htb]
	\tiny
	\label{table11}
	\centering
	\resizebox{\textwidth}{!}{
		\begin{tabular}{| p{5cm} | p{5cm} | }
			\multicolumn{2}{c}{\textbf{PLDRv51G-106M-2 with KV-cache enabled}} \\
			\hline 
			\textbf{Input} & \textbf{Continuation} \\ 
			\hline
			I love sci-fi and am willing to put up with a lot. Sci-fi movies/TV are usually underfunded, under-appreciated and misunderstood. I tried to like this, I really did, but it is to good TV sci-fi as Babylon 5 is to Star Trek (the original). Silly prosthetics, cheap cardboard sets, stilted dialogues, CG that doesn't match the background, and painfully one-dimensional characters cannot be overcome with a 'sci-fi' setting. What I would like to say is 
			&
			that the ex-biplo's story is pretty much all set on the spectrum. Maybe a little movie, or a TV sci-fi animated movie. At the end of the day, it's about the grit and determination of character and everything. On the show, the set is awful, and you have to make certain you don't throw out any much (not that many sci-fi series would have ever been realistic). That's not to say that you don't get good scores in 'Reckless Red' and 'Bricky Red'. It's an amazing live-action, forgettable set, with fantastic acting, an interesting script, and more. You can actually keep it on your game. It is quite a complicated story, but it is really exciting and interesting. There are really good, but not much of a comedic film. This one is something I like to consider. It's a very serious story about great characters, bigotry, and big heart. It is really a thoroughly enjoyable script, and a great challenge for the people that are trying to get the best out of it. Another great episode is some heavy steel, there are a few monsters and a few trucks and airplanes. Also
			\\
			\hline
	\end{tabular} } 
\end{table}

\begin{table}[!htb]
	\tiny
	\label{table12}
	\centering
	\resizebox{\textwidth}{!}{
		\begin{tabular}{| p{5cm} | p{5cm} | }
			\multicolumn{2}{c}{\textbf{PLDRv51G-106M-3 with KV-cache enabled}} \\
			\hline 
			\textbf{Input} & \textbf{Continuation} \\ 
			\hline
			I love sci-fi and am willing to put up with a lot. Sci-fi movies/TV are usually underfunded, under-appreciated and misunderstood. I tried to like this, I really did, but it is to good TV sci-fi as Babylon 5 is to Star Trek (the original). Silly prosthetics, cheap cardboard sets, stilted dialogues, CG that doesn't match the background, and painfully one-dimensional characters cannot be overcome with a 'sci-fi' setting. What I would like to say is 
			&
			that it does have some repetitive, frustrating ending to it. My interpretation of this story was something that is still right on the Internet. On the other hand, it might be something that you are very passionate about. I have always struggled with humor and was fascinated by it. I also have a couple of extra things that I love to see done, and one of them is looking at the world that is becoming a leader and then pushing people forward to make new choices. There is a part of me that goes in that my sceptical part is that it is a different medium. The implication is that it is not a popular medium. I still believe that it is best for me to have more of a long, self-reliant career, but also to be competitive. The eponymous meaning of the wording is that it has become an adversarial part of me. The absence of the actual wording is simply too many to bear. With words that are inherently human, the title of the piece is a pretty key part of me. For me, I don't like to seem good at being interested in other things. To not say that I do not like being upset with people is wrong. Rather, I have a way of
			\\
			\hline
	\end{tabular} } 
\end{table}

\end{document}